\documentclass[a4paper, 10pt, conference]{cssconf}
\IEEEoverridecommandlockouts    
\usepackage{epsfig} 
\usepackage{amsmath} 
\usepackage[bottom]{footmisc}
\usepackage[left=15mm,top=14mm,right=15mm,bottom=17mm]{geometry}
\title{\LARGE \bf
Isometric force pillow: using air pressure \\to quantify involuntary finger flexion in the presence of hypertonia
}

\author{Caitlyn E. Seim, Chuzhang Han, Alexis J. Lowber, Claire Brooks, Marie Payne, \\  \hspace{20 pt}  Maarten G. Lansberg, Kara E. Flavin, Julius P. A. Dewald, Allison M. Okamura
\thanks{This research was supported, in part, by the Stanford Wu Tsai Neurosciences Institute Neuroscience:Translate Program and the Eunice Kennedy Shriver National Institute Of Child Health and Human Development of the National Institutes of Health (NIH) under Award Number F32HD100104. 
}
\thanks{Author contact address is cseim@stanford.edu}
\thanks{C. E. Seim, C. Han, and A. M. Okamura are members of the Stanford University Department of Mechanical Engineering.  A. J. Lowber is with the Stanford University Department of Computer Science. M. Payne is part of the Stanford University Department of Civil Engineering. 
M. G. Lansberg and K. Flavin are with the Stanford University Department of Neurology.  J. P. A. Dewald is with the Northwestern University Department of Physical Therapy and Human Movement Sciences.}
}

\begin{document} 
 \setlength\dbltextfloatsep{0.277cm}
\maketitle
\thispagestyle{empty}
\pagestyle{empty}

\begin{abstract}
Survivors of central nervous system injury commonly present with spastic hypertonia.  The affected muscles are hyperexcitable and can display involuntary static muscle tone and an exaggerated stretch reflex. These symptoms affect posture and disrupt activities of daily living.  Symptoms are typically measured using subjective manual tests such as the Modified Ashworth Scale; however, more quantitative measures are necessary to evaluate potential treatments.  
The hands are one of the most common targets for intervention, but few investigators attempt to quantify symptoms of spastic hypertonia affecting the fingers.  
We present the isometric force pillow (IFP) to quantify involuntary grip force.  This lightweight, computerized tool provides a holistic measure of finger flexion force and can be used in various orientations for clinical testing and to measure the impact of assistive devices. 

\end{abstract}

\section{BACKGROUND}
Survivors of central nervous system injury often present with spastic hypertonia (involuntary muscle tone and spasticity).   
Abnormal supraspinal drive post-injury leads to motoneuron hyperexcitability  \cite{mcpherson2018altered}, resulting in involuntary muscle contractions that can impact posture and limit range of motion.  
During passive movement, the limb can show an exaggerated stretch reflex  
as a function of movement velocity (spasticity). 
At rest, the limb is commonly bent or flexed from static muscle tone (hypertonia).  
When the upper limb is affected, symptoms may cause pain and prevent activities of daily living such as hand washing and dressing.   

A variety of treatment options are used clinically, including splints,  pharmaceuticals, and surgical interventions.  
More treatments are also in development, such as therapeutic stimulation and robotic devices. 
The quantification of symptoms is both clinically and scientifically relevant.  However, 
most therapists and investigators still rely on manual ratings that can be highly subjective  
\cite{caliandro2012focal, mcgibbon2013elbow, germanotta2020reliability, seth2015robotic}. 
The most common measure, the Modified Ashworth Scale (MAS), is 
known to be subjective and have poor intra- and inter-rater variability \cite{yam2006interrater}. 
Other measures like the Modified Tardieu Scale (MTS) have also been shown to have high variability~\cite{centen2017kaps}. 
 
Some tools have been developed to quantitatively assess symptoms, but much of this work focuses on the limb's response to imposed movement.  
The affected muscle is stretched at constant velocities by a therapist or machine while joint angle is recorded using robotic devices  
\cite{seth2015robotic}, fiber optic tools \cite{mcgibbon2013elbow}, or marker tracking.  
Simultaneously, stiffness is measured through EMG \cite{mcgibbon2013elbow, alhusaini2010evaluation}, torque sensors \cite{germanotta2020reliability, seth2015robotic, centen2017kaps}, or load cells \cite{alhusaini2010evaluation}.  
Although such tools that move the limb 
can provide data on spasticity, imposed movement is relatively uncommon in daily life.  Since the limb is often at rest, measurement of static symptoms (such as contorted posture or involuntary muscle torque) could provide useful data
.  
Few existing tools measure static symptoms of hypertonia
. Germanotta et al. used a large robotic device  \cite{germanotta2020reliability}, and Stienen et al. created a 2-DoF hinge for the wrist and MCP joints~\cite{stienen2011wrist}. 

Most work on quantifying spastic hypertonia also focuses on the elbow, knee, or ankle joints \cite{mcgibbon2013elbow, seth2015robotic, centen2017kaps, alhusaini2010evaluation};  
yet the hands are one of the most common areas for intervention \cite{ren2009developing}.  
Hand function is key to performing many tasks of self sufficiency, and unchecked hypertonia in the hands can lead to secondary problems \cite{heijnen2008long}. 
Thus, measuring the level of spastic hypertonia on the hands is particularly important.  
However, the fingers are particularly difficult to analyze due to their many degrees of freedom.  Actuating each finger or all the phalanges together is challenging due to different phalanx bone lengths.  The fingers can also be difficult to secure to a device in the presence of hypertonia. 
Here we present the isometric force pillow (IFP): a lightweight, handheld tool designed to provide a holistic, quantitative measure of involuntary hand flexion due to spastic hypertonia.

\begin{figure*}
\centering
\includegraphics[width=\linewidth]{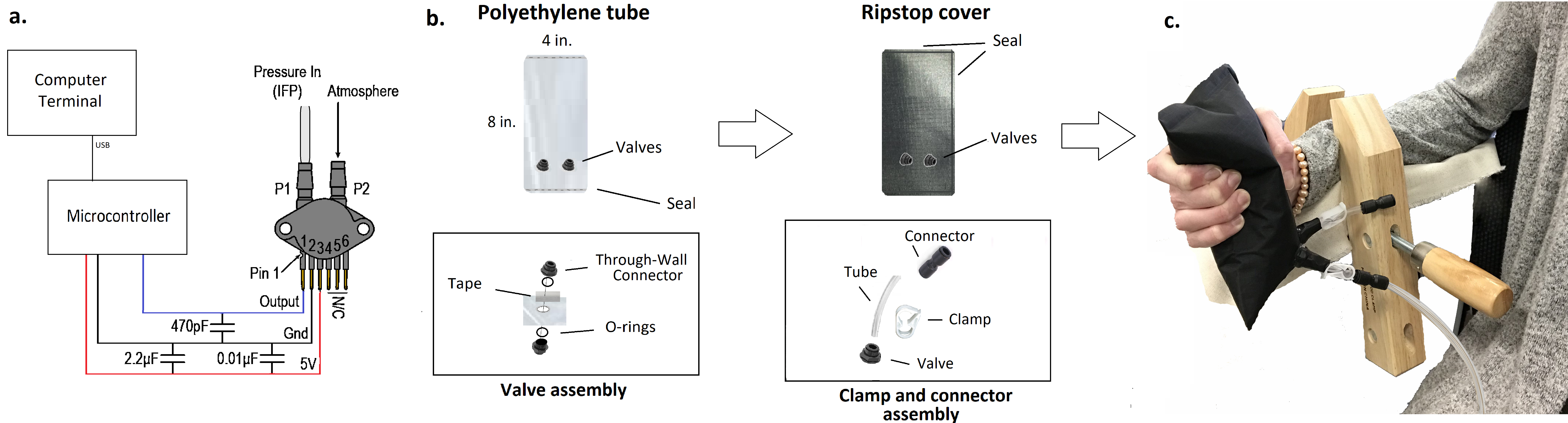}
 \caption{\textbf{a.} Schematic of the IFP electronics. 
\textbf{b.}  Assembly of the IFP.  
 \textbf{c.}  The isometric force pillow used in a gravity-neutral position.  The tube leads to the pressure sensor of \ref{image}a. The other valve is clamped after initial inflation using a hand vacuum pump.   
 }
\label{image}
\end{figure*}

\section{DESIGN} 
The objective was to develop a tool to measure symptoms of spastic hypertonia in the hands, in order to provide quantitative 
data when studying treatment efficacy. 
Spastic hypertonia often causes the finger flexors to contract, so measuring flexion force is an accepted strategy~\cite{germanotta2020reliability, stienen2011wrist, lan2017impact}.

The tool was intended to be low-cost and compact in order to promote perceived ease of use. Preliminary prototypes used a hinge-like design that measured finger flexion using a load cell -- aiming to expand the Wrist Finger Torque Sensor \cite{stienen2011wrist} into a stand-alone design with a focus on the fingers.  
When the preliminary prototypes were tested on stroke survivors with upper-limb hypertonia, the fingers were difficult to secure even using straps.  Like other tools \cite{germanotta2020reliability, seth2015robotic, stienen2011wrist, alhusaini2010evaluation} the hinge had to be mounted on a rigid surface. If the individual's elbow or wrist was contracted, their hand could not reach the tool.  Those with moderate to severe spastic hypertonia could not use the prototypes.  This design was also only capable of measuring force at the MCP joints. 

The hinge design was replaced with a graspable cushion (Fig. 1c), the isometric force pillow (IFP).    
The IFP has an ergonomic form similar to some orthotic devices for hypertonia, such as the Hand Contracture Carrot Orthosis (AliMed, Inc.). 
The IFP uses air pressure to measure finger flexion force; which provides a holistic measure of contraction at multiple joints. This self-contained method of measuring force does not require attachment to a rigid mount 
and thus allows the unit to be free-moving; it is capable of being used both in the gravity-neutral position and in other orientations.  The IFP is tethered only by a flexible tube. 

\subsection{Mechanical Design}
The cushion is 8 inches in length by 4 inches in diameter when not inflated. 
These dimensions allow fingers of various lengths to be extended when grasping the device, without allowing the thumb and fingertips to touch. 
Air pressure in the IFP is maintained to provide an accurate differential pressure measurement. An airtight seal is created in an \\8-step process (Fig. \ref{image}b).  

4-mil heavy duty polyethylene tubing cut at a length of 8 inches forms the inner layer of the cushion.  
Two air-tight valves are assembled using through-wall connectors (5779K677, McMaster-Carr).  Holes of 1/4'' diameter are cut in the polyethylene tubing and double-sided VHB acrylic tape (S-10123, Uline) is applied surrounding each hole.  
1/4'' rubber O-rings are placed on both the inside and the outside of the polyethylene tubing, and the through-wall connectors are twisted shut. 
After the valves are in place, a tabletop impulse sealer (H-163, Uline) seals the polyethylene tubing. 

A sheet of silicone-coated ripstop nylon (FRCS, Seattle Fabrics Inc.) is cut to wrap over the polyethylene tubing and is sealed on by the tabletop impulse sealer. The nylon material adds friction to prevent slippage while the IFP is in use. It also prevents the polyethylene tubing from stretching during inflation and squeezing, and prevents bursting. Four inches of 1/4'' soft plastic tubing attaches to each valve and a mini tubing clamp (59199, U.S. Plastic Corp.) is added so each valve could be sealed.  
One of these tubes attaches to a pressure sensor using a straight tube connector (5779K14, McMaster-Carr).  The other tube is used to inflate the cushion via a hand vacuum pump (MV8255, Mityvac). After inflation, the pump is disconnected and the tube is clamped.

\subsection{Electronics and Software}
A 50kPa differential pressure sensor (MPX5050-DP, NXP Semiconductors) measures pressure within the IFP.
One port (P1) on the sensor connects to the IFP valve via soft tubing.  The other port (P2) is exposed to atmospheric pressure. The power, ground, and output voltage pins on the sensor connect to a microcontroller (Arduino Duemilanove) using a 3-wire ribbon cable. The circuit board houses a power supply decoupling and output filtering circuit (Fig. 1a). 
 
A custom script converts the analog voltage from the sensor into gauge pressure readings at 10 Hz. The sensor readings are calibrated by subtracting a constant offset. The offset was empirically measured with both sensor P1 and P2 exposed to atmospheric pressure. The calibrated pressure values are smoothed using a moving average filter with a sample size of 10. 
The pressure reading is then displayed on a computer terminal for clinicians or investigators. 

\section{DISCUSSION}
The IFP tool provides a holistic, quantitative measure of involuntary grip due to spastic hypertonia. 
The majority of other tools that aim to quantify symptoms of spastic hypertonia impose movement on the limb, which provides data primarily on spasticity.  In contrast, the IFP measures static flexion force from the fingers.  Since static flexion leads to many secondary problems in spastic hypertonia, our tool provides valuable data that might also indicate problems such as difficulty to access and clean the palm and the progression of contractures. 
The tool could also be used to evaluate involuntary grip when using
assistive devices or pharmacological interventions to relieve hypertonia. 
  
Circumstantial factors must always be controlled when measuring muscle tone and spasticity.   Arm position is one such factor \cite{mcpherson2018altered} and measurement using the IFP in a standardized, gravity-neutral position is optimal.  Those with difficulty achieving a standardized arm position due to severe hypertonia can get repeated measures using the IFP to provide intrasubject trends. 
Most prior work relies on mounted hardware that is not accessible to some patients \cite{germanotta2020reliability, seth2015robotic}. 
For other patients it can be necessary to repeatedly stretch the affected limb to fit into these tools, and these stretches impact the validity of measures by temporarily reducing hypertonia~\cite{schmit2000stretch}.

\bibliographystyle{bmc-mathphys}  
\bibliography{bmc_article}       

\end{document}